\documentclass[11pt]{amsart}
\usepackage[margin=25mm]{geometry}
\usepackage{amsaddr} 
\usepackage{graphicx}
\usepackage{pgfplots}
\usepackage{tikz}
\usepackage{tikzscale} 
\usepackage{hyperref}
\usepackage[margin=25mm]{geometry}
\usepackage{booktabs}
\usepackage{setspace}
\pgfplotsset{compat=1.16}
\usetikzlibrary{arrows.meta}
\usetikzlibrary{backgrounds}
\usepgfplotslibrary{patchplots}
\usepgfplotslibrary{fillbetween}
\pgfplotsset{%
    layers/standard/.define layer set={%
        background,axis background,axis grid,axis ticks,axis lines,axis tick labels,pre main,main,axis descriptions,axis foreground%
    }{
        grid style={/pgfplots/on layer=axis grid},%
        tick style={/pgfplots/on layer=axis ticks},%
        axis line style={/pgfplots/on layer=axis lines},%
        label style={/pgfplots/on layer=axis descriptions},%
        legend style={/pgfplots/on layer=axis descriptions},%
        title style={/pgfplots/on layer=axis descriptions},%
        colorbar style={/pgfplots/on layer=axis descriptions},%
        ticklabel style={/pgfplots/on layer=axis tick labels},%
        axis background@ style={/pgfplots/on layer=axis background},%
        3d box foreground style={/pgfplots/on layer=axis foreground},%
    },
}

\begin{document}
\bibliographystyle{plos2015}
\title%
[Comparing tumor growth models]%
{New tools for comparing classical and\\%
  neural ODE models for tumor growth}

\author{Anthony D.~Blaom}
\address{Department of Computer Science\\
  University of Auckland\\
  New Zealand}
\author{Samuel Okon}
\address{German Research Center for Artificial Intelligence\\
  Kaiserslautern\\
  Germany}
\email{anthony.blaom@gmail.com, samuel.okon@dfki.de}
\thispagestyle{empty}
\begin{abstract}
  A new computational tool {\ttfamily TumorGrowth.jl} for modeling tumor growth is
  introduced. The tool allows the comparison of standard textbook models, such as General
  Bertalanffy and Gompertz, with some newer models, including, for the first time, neural
  ODE models. As an application, we revisit a human meta-study of non-small cell lung
  cancer and bladder cancer lesions, in patients undergoing two different treatment
  options, to determine if previously reported performance differences are statistically
  significant, and if newer, more complex models perform any better. In a population of
  examples with at least four time-volume measurements available for calibration, and
  an average of about 6.3, our main conclusion is that the General Bertalanffy model has
  superior performance, on average. However, where more measurements are available, we
  argue that more complex models, capable of capturing rebound and relapse behavior, may
  be better choices.
\end{abstract}
\maketitle


\section{Introduction}
We investigate the performance of models for the growth of tumors, as measured by a single
parameter, typically the volume. Models under consideration are based on solving ordinary
differential equations (ODE's). These ODE's have unknown parameters, which typically means
forecasting a tumor's future size is only possible after calibrating the model using the
current clinical history. We introduce a new package, {\ttfamily TumorGrowth.jl}
\cite{Blaom_2024}, to automate this procedure, for a battery of classical models, such the
General Bertalanffy model \cite{Kuang_etal_2016,Norton_etal_76}, as well more complex
models, which include, for the first time, neural ODE's \cite{Chen_etal_2018}. Custom
models can also be implemented.

\subsection{Previous evaluations of classical model performance}
As an application of the new tool, we revisit the study of Laleh et
al.~\cite{Laleh_etal_2022}, the first of its kind, which includes an out-of-sample
evaluation of the accuracy of classical ``textbook'' models in a meta-study of 652 tumors,
in humans undergoing chemotherapy or cancer immunotherapy. Patients in that study are
either non-small cell lung cancer or bladder cancer sufferers. The specific treatments
compared are Atezolizumab (previously known as MPDL3280A) and Docetaxel. Refer to
\cite{Laleh_etal_2022} for details. Models in the meta-study are ranked based on the mean
absolute error on a holdout test set. The models compared are: exponential, logistic,
classical Bertalanffy, General Bertalanffy, classical Gompertz, and the General Gompertz
models. Each study and study arm gets a separate treatment in \cite{Laleh_etal_2022} and
aggregated scores are not reported. However the conclusion is a general trend favoring the
General Bertalanffy and the classical Gompertz models.

That said, no statistical significance is attached to these results. It could be that, by
chance, the actual expected performance of these models is different from the reported
performance, and the likelihood of this scenario remains unquantified in the study. The
present study provides the missing statistical analysis, and adds some newer models to the
comparison.

\subsection{Capturing relapse or rebound behavior}

\begin{figure}[h]
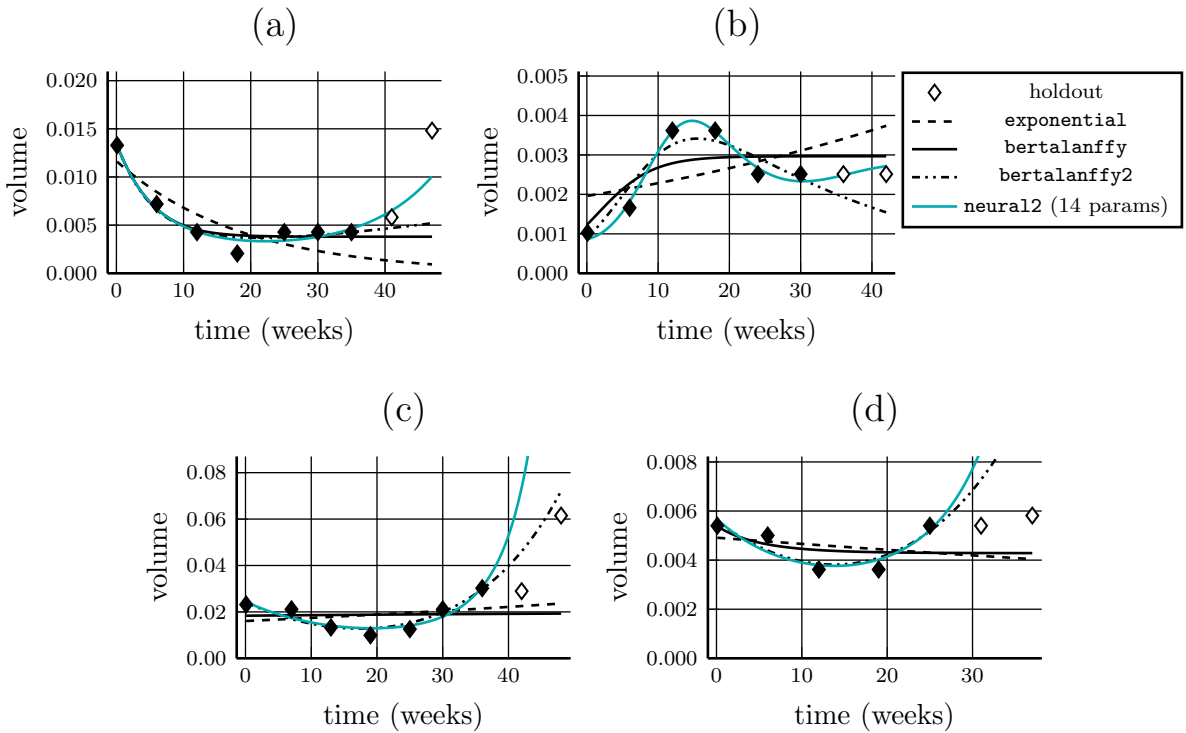

  \begin{center}
     \includegraphics{4.tikz}\includegraphics{12.tikz}\\
     \includegraphics{14.tikz}\includegraphics{1.tikz}\\
   \end{center}
   \caption{Selected model comparisons for observations with relapse or rebound. Solid
     diamonds indicate calibration data, open diamonds subsequent observations not used in
     calibration. Two classical models, exponential and General Bertalanffy ({\ttfamily
       bertalanffy}), are compared with a 2D generalization of General Bertalanffy
     ({\ttfamily bertalanffy2}) and a 2D, 14-parameter neural ODE ({\ttfamily
       neural2}). While the newer models perform better on the holdout data in (a), the
     classical models do better in (d). Results are mixed in cases (b) and (c).}
   \label{relapse}
\end{figure}

As they are first order, one-dimensional ODE's, none of the classical models in the Laleh
et al.~ study can capture rebound or relapse phenomena: Solutions are always monotonically
increasing, decreasing or constant. This is not surprising, as the classical models were
constructed for untreated lesions.  In the current study (and in {\ttfamily
  TumorGrowth.jl}) we consider two models in which the volume is coupled to a second
latent variable (making them effectively second-order ODE's) and which do not suffer this
limitation:
\begin{itemize}
\item A novel but simple 2D generalization of the General Bertalanffy model described in
  \ref{bertalanffy2}, with one additional parameter (5 total, including initial condition)
\item A basic 2D neural ODE, with 14 parameters, described in \ref{neural}.
\end{itemize}

A {\slshape neural ODE} is an ODE $\dot x = F(x, \theta)$ where $F(x, \theta)$ represents
the output of an artificial neural network, with input $x$, and a system of internal
weights and biases $\theta $ \cite{Chen_etal_2018}. For simplicity, we focus, in our
comparisons, on a particular neural ODE, i.e., on a particular network architecture. Note,
however, that by varying the architecture (which is possible in {\ttfamily
  TumorGrowth.jl}) one can approximate any ODE arbitrarily well.  This follows from the
Universal Approximation Theorem for neural networks \cite{Hornick_etal_1989}. Although the
precise mechanisms of tumor growth or mitigation may be unclear, one tends to believe the
underlying process is nevertheless governed by {\em some} system of ordinary differential
equations. That is, one views the use of a neural ODE as more ``physically'' justified
than ordinary curve fitting with, say, polynomials.

As one can see from Figure 1(a), the two new models have the potential to make
better predictions in the presence of a rebound or relapse. A neural ODE model can capture
even more nuanced behavior, as demonstrated in (b). However, the danger of using models
with more free parameters is over-fitting, which is clearly evident in (c) (for the neural
ODE model) and (d) (for both new models). The present study tackles the question of which
tendency --- better prediction or over-fitting -- is the more typical in the clinical
context.

Obviously, answers to the question just posed depends on the number of observations
available for calibration. Following Laleh et al., we restrict attention to observations
in their meta-study exceeding six in number. Deviating slightly from Laleh et al., we hold
back the last two, rather than three, observations for testing. The average number of
observations in this restricted dataset, including hold-outs, is 8.3, with about 73\% less
than 9.

\subsection{Results}\label{results}
We now summarize the results of calibration experiments detailed in Section
\ref{study}. Our analysis treats the observations collated in the Laleh et al.~study as a
whole, with no attempt to stratify outcomes over individual study arms. We are doubtful
statistically significant conclusions can be drawn at the level of study arms, because of
small sample size, as it is already difficult in the aggregated case, as we now report.

Where there are differences in the studies, or in individual patients, about the timing of
treatment, these are also conflated in the present analysis.

Figure 2 shows a boxplot for the prediction errors for each model. No
statistically significant difference in model performance is evident in these
results. Although it has dubious probative value, we record in Table 1
rankings based on the mean absolute prediction errors.

\begin{figure}[h]\label{boxplots}
  \begin{center}
     \includegraphics{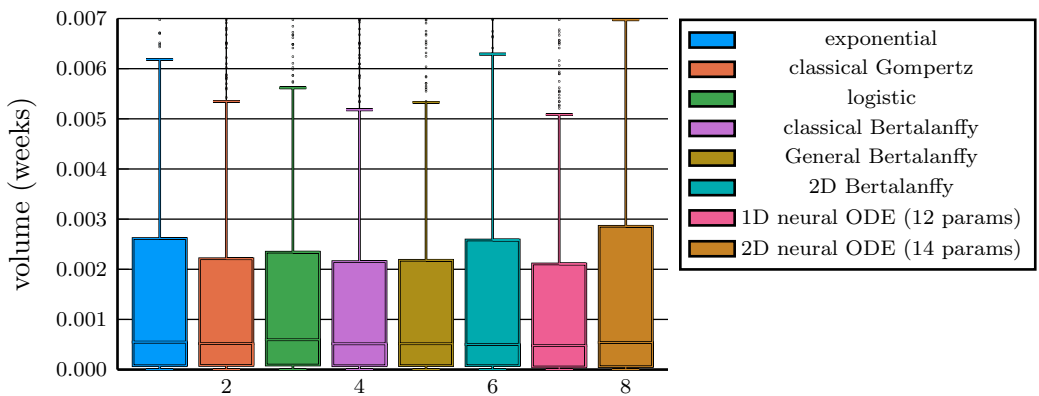}
   \end{center}
   \caption{Box-and-whisker plots of the absolute prediction errors on a holdout set.}
\end{figure}

{\small
\begin{table}[h]
    \centering
    \begin{tabular}{llll}\toprule
                 model  & mean abs. error & num.~parameters & section \\ \midrule
           General Bertalanffy   & 0.0027 & 4 & \ref{classical} \\
 classical Bertalanffy           & 0.0028 & 3 & \ref{classical} \\
              classical Gompertz & 0.0028 & 3 & \ref{classical} \\
              logistic           & 0.0029 & 3 & \ref{classical} \\
              1D neural ODE      & 0.0031 & 12 & \ref{classical} \\
          2D General Bertalanffy & 0.0032 & 5 & \ref{bertalanffy2} \\
           exponential           & 0.0033 & 2 & \ref{exponential} \\
               2D neural ODE     & 0.0046 & 14 & \ref{neural} \\ \bottomrule
    \end{tabular}
    \vspace{\baselineskip}
    \caption{Models ranked by mean absolute error on a two-observation holdout set, but note large standard errors shown in Figure 2.}
  \end{table}
}

Turning to a more powerful test, we compute 95\% bootstrap confidence intervals for the
{\em pairwise difference} in individual lesion prediction errors. When such an interval
contains zero, we declare a statistical tie, otherwise a win is declared for the better
performing model. The outcomes of these match-ups, presented in Table 2, suggest:

\begin{itemize}
\item The General Bertalanffy model ({\ttfamily bertalanffy}) is superior to all other
  models, except the 1D neural ODE ({\ttfamily neural}), where no statistically
  significant difference is detectable. However, there is no detectable difference between
  the 1D neural model and {\em any} other model, which is a fair basis for rejecting it as
  an alternative in the current context.
  \item The 14-parameter 2D neural ODE tested in this study is inferior to all other models.
  \item The exponential model is inferior to most models and not demonstrably superior
    to any other classical models.
  \item Otherwise, there are no statistically significant differences between the models.
\end{itemize}

{\small
\begin{table}[h]
    \centering
    \begin{tabular}{llllllll}
      \toprule
                  model &   c. Gomp.   & log.         &     c.~Bert. & G. Bert.   & 2D Bert.     & 1D neural & 2D neural \\ \midrule
            exponential &   $\uparrow$ &         draw &   $\uparrow$ & $\uparrow$ &         draw &      draw & $\leftarrow$ \\
     classical Gompertz &              &         draw &         draw & $\uparrow$ &         draw &      draw & $\leftarrow$ \\
               logistic &         draw &              &         draw & $\uparrow$ &         draw &      draw & $\leftarrow$ \\
  classical Bertalanffy &         draw &     draw     &              & $\uparrow$ &         draw &      draw & $\leftarrow$ \\
    General Bertalanffy & $\leftarrow$ & $\leftarrow$ & $\leftarrow$ &            & $\leftarrow$ &      draw & $\leftarrow$ \\
         2D Bertalanffy &         draw &         draw &         draw & $\uparrow$ &              &      draw & $\leftarrow$ \\
          1D neural ODE &         draw &         draw &         draw &       draw &         draw &           & $\leftarrow$ \\ \bottomrule
    \end{tabular}
    \vspace{\baselineskip}\label{model_battle}
    \caption{Head-to-head model comparisons. Where ``draw'' appears, the difference in
      model performance is not significant at the 5\% level. Where there is a significant
      difference, an arrow points to the superior model.}
\end{table}
}
\section{The {\ttfamily TumorGrowth.jl} package}
The {\ttfamily TumorGrowth.jl} package is a tool for calibrating and comparing models for
tumor growth, as measured by a single parameter, typically volume \cite{Blaom_2024}. It
features:

\begin{itemize}
\item Implementations of the classical and General Bertalanffy, classical Gompertz and
  logistic models (see \ref{classical} below) as well as the simple model for exponential
  decay or growth (\ref{exponential})
\item A two-dimensional generalization of General Bertalanffy with one extra parameter, for
  capturing rebound or relapse behavior (\ref{bertalanffy2})
\item One and two-dimensional neural ODE models
  (\ref{neural})
\item Sophisticated control over parameter optimization, such as early stopping criteria
\item An option to give higher weight to more recent measurements during calibration (not
  applied in the study above)
\item Comprehensive options for optimization, either by gradient descent,
  Levenberg-Marquardt, or Powell's dog leg algorithm \cite{Nocedal_Wright_06}
\item Plotting functions to visualize results
\item The ability to specify a custom model
\item Convenient access to data from the human meta-study collated in
  \cite{Laleh_etal_2022}.
\end{itemize}

For further details, and tutorials, the reader is referred to the comprehensive package
documentation \cite{Blaom_2024}.

{\ttfamily TumorGrowth.jl} is written in pure {\ttfamily Julia}, which has the advantages
of speed, high customizability, transparency and reproducibility.  Beyond the classical
case where analytic solutions to the underlying ODE's are known, model calibration
involves differentiating numerically obtained solutions with respect to parameters and
initial conditions. In this respect, development of {\ttfamily TumorGrowth.jl} was able to
capitalize significantly on Julia's state-of-the-art {\ttfamily SciML} ecosystem
\cite{Ma_etal_2021}, which implements automatic differentiation, and in particular,
Pontryagin's adjoint method for differentiating ODE solutions.

\section{Models}
In this section we provide the detailed specification of models tested in our study
reported above, and implemented in {\ttfamily TumorGrowth.jl}.

\subsection{The General Bertalanffy model}\label{classical}

In its common formulation, Bertallanfy's model for lesion growth
\cite{Bertalanffy_57,Kuang_etal_2016} is
\begin{equation}
    \frac{dv}{dt} = \eta v^m - \kappa v^n.\label{b}
\end{equation}
Here $v$ denotes lesion volume, $t$ is time; $\eta$, $\kappa$, while $m$ and $n$ are
parameters. Bertalanffy argued that solutions to \eqref{b} are relatively insensitive to
variations of $n$ close to unity. In this article, and following \cite{Laleh_etal_2022},
we restrict to the special $n=1$ case, giving what will be called the {\slshape General
  Bertalanffy} model\footnote{ {\ttfamily bertalanffy} in {\ttfamily TumorGrowth.jl}}. In
this case, Bertalanffy was able to provide an analytic solution, reproduced below. Further
specializing to the case $m=2/3$, as Bertalanffy did, we obtain the {\slshape classical
  Bertalanffy} model\footnote{{\ttfamily classical\_bertalanffy}}. As we shortly recall,
the classical logistic or Verhulst\footnote{{\ttfamily logistic}} and
Gompertz\footnote{{\ttfamily gompertz}} models for tumor growth
\cite{Kuang_etal_2016,Norton_etal_76} can also be regarded as special cases of the General
Bertalanffy model. However, instead of using Equation \eqref{b}, we prefer a reformulation
with new parameters that pays attention to dimensional correctness, described next.

Recall that the Box-Cox transformation with parameter $\lambda $ is defined by
\begin{equation*}
  B_\lambda(x) = \begin{cases}
  \frac{x^\lambda - 1}{\lambda }  & \text{if $\lambda\ne 0$} \\
  \log(x) & \text{if $\lambda = 0$}
\end{cases}.
\end{equation*}
This function is not only continuous at $\lambda = 0$, but is an infinitely differentiable
function of $x$ and $\lambda $, defined for all real $\lambda $ and $x > 0$. If $n=1$,
then, after a change of parameters, \eqref{b} is equivalent to
\begin{equation}
  \frac{dv}{dt} = \omega B_\lambda\left(\frac{v_\infty}{v}\right)v, \qquad v>0,\label{bert}
\end{equation}
as is readily established. Here $\omega $ is a parameter with the units of inverse time,
and $v_\infty>0$ a parameter with the units of volume.  The non-dimensional Box-Cox
exponent $\lambda $ has the following interpretation:
\begin{itemize}
  \item If $\lambda = 1/3$, \eqref{bert} is the classical Bertalanffy model (i.e, with $m=2/3$ in \eqref{b}).
  \item If $\lambda = -1$,  \eqref{bert} is the logistic model.
  \item If $\lambda =0$,  \eqref{bert} is the classical Gompertz model.
\end{itemize}

Supposing $\omega>0$, the qualitative dynamics of \eqref{bert} is always the same. There
is a single steady state solution $v(t) = v_\infty$, which is always stable. As a first
order, autonomous ODE in one dimension, all its other solutions are either monotonically
increasing (true here if $v(0) < v_\infty$) or monotonically decreasing (true if
$v(0) > v_\infty$).  Writing $v_0=v(0)$, Bertalanffy's analytic solution is
\begin{equation*}
     v(t) =
     \begin{cases}
       \left(1 + \left(\left(\frac{v_0}{v_\infty}\right)^\lambda - 1\right)
         e^{-\omega t}\right)^{1/\lambda}v_\infty & \text{if $\lambda \ne 0$}\\
       \left(\frac{v_0}{v_\infty}\right)^{e^{-\omega t}}v_\infty & \text{if $\lambda = 0$}
     \end{cases}.
\end{equation*}
If we non-dimensionalize volumes by $v_\infty$ and times by $1/\omega $, then the
solutions can be written
\begin{equation*}
     v(t) =
     \begin{cases}
       \left(1 + \left(v_0^\lambda - 1\right)
         e^{-t}\right)^{1/\lambda}& \text{if $\lambda \ne 0$}\\
       v_0^{e^{-t}} & \text{if $\lambda = 0$}
     \end{cases}.
\end{equation*}

\subsection{A two-dimensional generalization of the General Bertalanffy
  model}\label{bertalanffy2}
In patients undergoing treatment, the growth of lesions is frequently not monotonic. For
example, a lesion may initially decrease in size, but ultimately begin increasing
again. To capture such behavior in an autonomous ODE we must adopt a higher order model
or, equivalently, a first order ODE with more than one dimension. A simple two-dimensional
extension\footnote{{\ttfamily bertalanffy2} in {\ttfamily TumorGrowth.jl}} of the
General Bertalanffy model \eqref{bert} is obtained by replacing the parameter
$v_\infty$ with a new latent variable $u(t)$, the (time-varying) {\slshape carrying
  capacity}, which we allow to evolve independently of $v(t)$, at a rate in proportion to
its magnitude:
\begin{equation}
  \frac{dv}{dt} = \omega B_\lambda\left(\frac{u}{v}\right)v, \qquad
  \frac{du}{dt} = \gamma \omega u.\label{bert2}
\end{equation}
Here $\gamma $ is a new dimensionless parameter, taking any real value, which introduces a
second time scale $1 \over |\gamma \omega|$. Since, $u$ is latent, the value of $u(0)$ is
unknown. Recycling notation, we retain $v_\infty$ as a model hyperparameter, but it is now
the initial carrying capacity $u(0)$. Then, taking $\gamma =0$, we get $u(t) = v_\infty$
for all $t$ , recovering the solution $v(t)$ to the first order model \eqref{bert}.

Our extension described by Equation \eqref{bert2} is it is not based on any physiological
mechanism known to us, but is one of the simplest ways to generalize \eqref{bert}.

\subsection{Neural ODE's}\label{neural}
A {\slshape neural ODE} is an ordinary differential equation of the form
$\dot x = F(x, \theta) $, where $F(x, \theta)$ is the output of some neural network with
input $x$, and some system of weights and biases $\theta $ \cite{Chen_etal_2018}. We
consider two classes of neural ODE models in this article, both implemented in {\ttfamily
  TumorGrowth.jl}. In the one-dimensional neural ODE\footnote{{\ttfamily neural}} a volume
scale $v_\infty$ and invertible transform
$\phi\colon (0, \infty) \rightarrow {\mathbb R} $ are specified, and we declare that the
non-dimensionalized, transformed volume $y(t) := \phi(v(t)/v_\infty)$ is to evolve
according to the ODE
\begin{equation*}
      \frac{dy}{dt} =f(y(t), \theta).
\end{equation*}
Here $f(\,\cdot\,,\theta)$ is any single input, single output, neural network, as
constructed using the {\ttfamily Lux.jl} framework, parameterized by $\theta$
\cite{Pal_2023a}. Typically, $\phi=\log$. In our two-dimensional neural
ODE\footnote{{\ttfamily neural2}}, $y(t)$ is defined as for the one-dimensional case, but
its evolution is coupled with that of a new latent variable $u(t) \in {\mathbb R}$:
\begin{align*}
  \frac{dy}{dt} &= f_1(y, u)\\
  \frac{du}{dt} &= f_2(y, u).
\end{align*}
Here, $f_1$ and $f_2$ are the components of any {\ttfamily Lux.jl} neural network
$f(\,\cdot\,, \theta)$ with two-dimensional input and output.

\subsection{Exponential decay or growth}\label{exponential}
For completeness, we also consider the simple model
$dv/dt = - \omega v$, whose solutions are exponential decay or growth.\footnote{{\ttfamily
    exponential}}

\section{Comparing the models}\label{study}
We now detail the computations leading to the results reported in \ref{results}. For full
details of the computation, refer to the ``Modal Battle'' section of the {\ttfamily
  TumorGrowth.jl} documentation \cite{Blaom_2024}.

A total of 652 lesion time series were extracted from the data collated in
\cite{Laleh_etal_2022}, by discarding any example with less than six measurements, which
leaves examples with an average of 8.3 measurements per lesion. These lesions come from
distinct patients undergoing chemotherapy or cancer immunotherapy. As detailed below, each
of the models listed in Table 1 was calibrated individually using the 652 examples, using
all but the last two measurements from each example. Then each calibrated model was used
to predict volumes for the two holdout times, and the average mean absolute deviation from
the measured volumes was recorded. These deviations were analyzed for statistical
significance, as already described in \ref{results}.

As in \cite{Laleh_etal_2022}, calibration is achieved by choosing the initial condition
$v_0$ and model parameters minimizing the sum of squares loss for the training
observations (not the mean absolute error). Instead of using the Trust Region Reflective
algorithm implemented in Python's {\ttfamily scipy} package (a variation on
Levenberg-Marquardt optimization) we used Adam gradient descent. {\ttfamily
  TumorGrowth.jl} also provides Levenberg-Marquardt optimization, but will not work for
the neural ODE models tested here, where the number of parameters to be optimized exceeds
the number of time-volume pairs.

\subsection{Addressing instability during calibration}
The parameter constraints $v_0, v_\infty > 0$ posed some difficulties, especially in the
classical models that have a singularity at zero volume. These issues persist if one
instead uses Levenberg-Marquardt or Powell's dog leg optimizers. A first step in
mitigating the issue was to arrange that {\ttfamily TumorGrowth.jl} handle parameter
bounds in the following (non-standard) way: If an optimization step leads to a parameter
moving out of bounds, then instead of moving immediately to the boundary (possibly a
singularity) the parameter steps half way towards the boundary. As a second
mitigation measure we were led to add a loss penalty discouraging large differences
between $v_0$ and $v_\infty$ on a log scale. It has the following form:
\begin{equation*}
  \text{penalized loss}\quad=\quad \left(\frac{v_0^2 + v_\infty^2}{2v_0v_\infty}\right)^\kappa
  \enspace\times\enspace\text{least squares loss},
\end{equation*}
where $\kappa$ is the penalty strength ({\ttfamily penalty} in {\ttfamily
  TumorGrowth.jl}).  Experiments led to a default of $\kappa = 0.3$ for the neural network
models, and $\kappa =0.8$ for the other models. Nevertheless, about 2.5\% of examples were
discarded due to instability issues, mostly associated with the neural ODE models.

Since the computations were not overly long, small learning rates were adopted, and
optimizers were run for many thousands of iterations, guaranteeing a high degree of
convergence to some local optimum. There was no guarantee that local optima were actually
global, which may have led to differences in some calibrations between the two studies.

\section{Discussion}
The {\ttfamily TumorGrowth.jl} package provides fast and convenient tools for calibrating
and comparing common ``textbook'' tumor growth models, as well as newer models, such as
neural ODE models.

The main conclusion from a statistical analysis of human treatment data, as first collated
and analyzed in \cite{Laleh_etal_2022}, is that the superior performance of the General
Bertalanffy model, when compared to the other models in Table 1, is statistically
significant, except in the case of the one-dimensional neural ODE model, which however
does not outperform any model to a statistically detectable degree. The exponential model
is the poorest classical model, and the two-dimensional neural ODE model is outperformed by
all other models. Otherwise, performance differences between the models are not
statistically detectable. It is to be emphasized that these conclusions apply only to
performance in the average, for a population of examples including at least 4 calibration
measurements, 72\% of which have less than 7 measurements. Transferring these conclusions
to populations with different characteristics is ill advised.

Where more measurements are available, more complex models, such as those which can
account for rebound or relapse behavior, are expected to outperform the classical
models. This is already evident in particular cases, as demonstrated by Figure 1. Possible
candidates include our two-dimensional generalization of the General Bertalanffy model,
and neural ODE models, provided these have a relatively a small of nodes to prevent
over-fitting. As more clinical data becomes available, finer, statistically detectable
differences in model performance may also be possible.

No attempt to stratify model comparisons over individual cancer types (lung versus
bladder) or over treatment types has been attempted here, as we suspect statistically
significant results will be elusive without more patient data, but this is another
interesting area for further investigation. An interesting exercise also not pursued here
is to analyze non-human data, where more statistically significant conclusions may be
possible.

An option for improving model performance, provided by the {\ttfamily TumorGrowth.jl}
package, is to give greater weight to more recent examples during calibration, and such
improvement was indeed observed in individual examples. However, introducing this option
essentially adds model complexity, as the degree of weighting arguably needs to be learned
along with model parameters; this increases the danger of over-fitting. For this reason,
the option was not fully investigated in the present study (and is not considered in
\cite{Laleh_etal_2022}) but it may be valuable in other contexts.

\subsection*{Acknowledgements}
The authors Sebastian J.~Vollmer for suggesting this investigation and Mark Gahegan for
detailed feedback on the manuscript. Help in structuring the meta-study data of
\cite{Laleh_etal_2022} was provided by Yasin Elmaci. The authors acknowledge useful
discussions with with Elmaci on unpublished work seeking to reproduce the results of
\cite{Laleh_etal_2022}.

\subsection*{Financial disclosures} The project on which this report is based was funded
by the German Federal Ministry of Education and Research under grant number
01IW23005. Responsibility for the content of this publication lies with the author.

\subsection*{Author Contributions}
Conceptualization: A.~B., S.~O.;
Data curation: S.~O.;
Investigation: A.~B.;
Methodology: A.~B., S.~O.;
Software development and documentation: A.~B.;
Numerical Experiments: A.~B.;
Writing – original draft: A.~B.;
Writing – review \& editing: A.~B., S.~O.

\end{document}